\documentclass{article} 
\usepackage{iclr2027_conference,times}
\usepackage{helvet}   

\usepackage{xcolor}
\definecolor{posdelta}{RGB}{0,128,0}
\definecolor{negdelta}{RGB}{180,0,0}

\usepackage{amsmath,amssymb}
\usepackage{algorithm}
\usepackage{algpseudocode}     
\usepackage{bm}
\usepackage{booktabs}
\usepackage{enumitem}
\usepackage{multirow}
\usepackage{graphicx}
\usepackage{iftex}
\ifXeTeX
  \usepackage[OT1]{fontenc}
  \usepackage[protrusion=true,expansion=false]{microtype}
\else
  \usepackage[protrusion=true,expansion=true,stretch=10,shrink=10,step=1]{microtype}
\fi
\usepackage{hyperref}
\usepackage{url}
\usepackage{fontawesome5}

\title{PhysEvolve: Verifier-Grounded\\
Self-Evolution for Physical Reasoning\\
with Executable Worlds}

\author{%
  Jinyuan Hu\thanks{Equal contribution.} \And
  Yilei Jiang\footnotemark[1] \And
  Chen Tang \And
  Minghong Cai \And
  Xinran Liu \And
  Xiangyu Yue\thanks{Corresponding author: \texttt{xyyue@ie.cuhk.edu.hk}.}%
}

\DeclareRobustCommand{\physevolve}{{PhysEvolve}}
\newcommand{\posd}[1]{{\scriptsize\color{posdelta}{(+#1)}}}

\iclrfinalcopy 
\newsavebox{\jointcaptionbox}
\newsavebox{\auditcaptionbox}
\newlength{\pairedcaptionheight}
\begin{document}

\maketitle
\lhead{Preprint.}
\vspace{-0.30in}
{\noindent\small
\hspace*{\tabcolsep}MMLab, The Chinese University of Hong Kong\\[3pt]
\hspace*{\tabcolsep}\faGlobe~Project page:~\url{https://yelonhu.github.io/physevolve-page/}\\[2pt]
\hspace*{\tabcolsep}\faGithub~GitHub:~\url{https://github.com/yelonhu/PhysEvolve}\par}
\vspace{0.05in}
\begin{abstract}
Physical reasoning underpins how embodied systems understand and act in the
physical world, yet vision-language models (VLMs) remain unreliable on
questions about dynamics and physical causality. Progress is critically
constrained by the scarcity of grounded counterfactual training data, as
observational videos capture only the factual trajectory that occurred. The
self-evolving paradigm offers a route beyond fixed datasets by enabling models
to construct their own training problems. Extending this paradigm to physics,
however, requires an external source of truth that can both validate
model-proposed environments and verify alternative physical outcomes under
intervention. We introduce \physevolve{}, a verifier-grounded self-evolving
framework that trains VLMs for causal physical reasoning from model-proposed
executable worlds and simulation-authored tasks. At each round, a shared VLM
proposes executable world programs and evidence schemas; a deterministic
physics engine freezes each valid world, simulates matched baseline and
body-elided counterfactual branches, and synthesizes discrete, answer-bound
visual tasks. Exact verifier feedback from task solving updates both the
shared policy via GRPO and an adaptive curriculum over physical intents.
Starting from Qwen3-VL-8B-Instruct, \physevolve{} achieves targeted gains across
PhysBench, CausalPhys, and ContPhy without human intervention data, improving
PhysBench Throwing by $8.8$ points and ContPhy Counterfactual Removal and
Goal-Pit by $7.4$ and $24.4$ points. These gains concentrate on
interaction-sensitive and altered-world questions, demonstrating zero-shot
transfer of causal principles to unseen complex dynamics (cloth, fluid, soft
bodies) and real-world video.
\end{abstract}

\section{Introduction}
\label{sec:intro}

\looseness=-1
Multimodal understanding of the physical world is foundational to embodied
intelligence and robotic systems, where reliable decision-making requires
anticipating how scenes evolve and respond to external interventions
\citep{black2024pi0,zhou2026paibench}. Yet current vision-language models
(VLMs) remain substantially less reliable on physical dynamics, intervention
effects, and counterfactual outcomes than on static object recognition or
spatial relations \citep{chow2025physbench,tang2026causalphys,zheng2024contphy},
and broader multimodal reasoning benchmarks echo the same reliability gap on
structured logical inputs \citep{yuan2025mmereasoning}.
Closing this gap is fundamentally bottlenecked by training data: high-quality,
verifiably annotated examples that isolate how physical outcomes change under
controlled interventions are exceptionally scarce. Real-world videos are not
only costly to annotate, but fundamentally capture only the factual trajectory
that occurred, rendering counterfactual alternatives unobservable
\citep{pearl2009causality}.

\looseness=-1
A growing paradigm of self-evolving systems circumvents static data bottlenecks
by enabling models to autonomously generate the problems from which they learn
\citep{zhao2025azr,xia2025agent0,gao2026survey}. The viability of this paradigm
hinges on an external source of truth that can independently validate
self-generated problems and verify their correct answers as the training
distribution shifts. In symbolic domains such as mathematics and code, an
interpreter provides this exact supervision: a Proposer generates an executable
specification, the interpreter computes its deterministic output, and exact
correctness supplies the Solver's learning signal. Physics simulators offer a
natural counterpart for the physical world by executing proposed environments
and exposing their complete internal states \citep{prabhudesai2026sim2reason}.
This analogy suggests an appealingly direct route to physical self-play: retain
the proposer--solver curriculum and substitute a physics simulator for the
symbolic interpreter.

\begin{figure}[!tbp]
  \centering
  \setlength{\abovecaptionskip}{5pt}
  \includegraphics[width=0.95\linewidth,trim=0 9 0 2,clip]{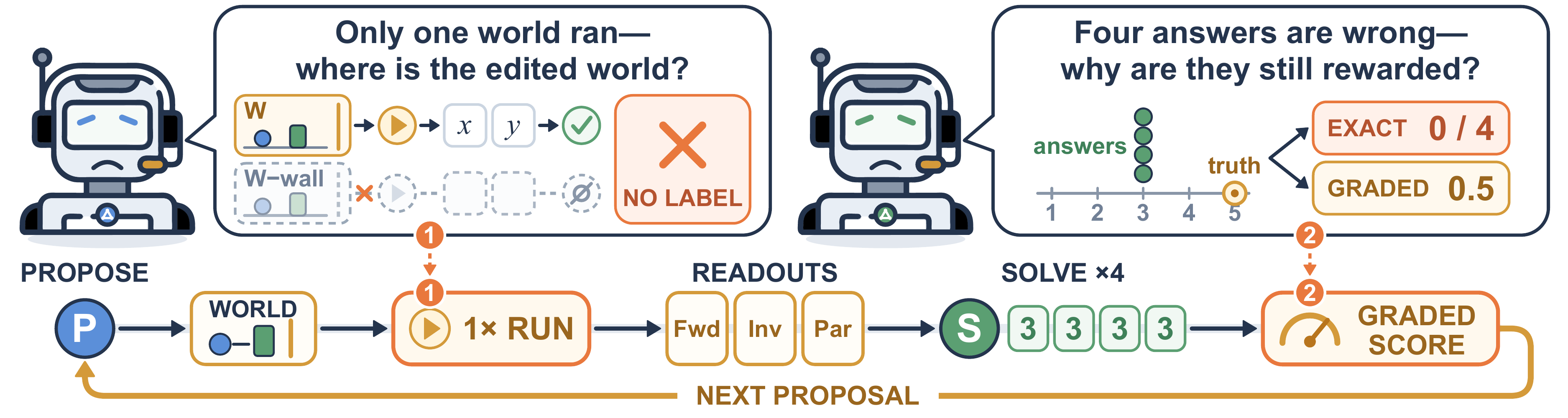}
  \caption{Naive self-play baseline for physical reasoning. A Proposer
  generates a world program; the executor simulates a single trajectory and
  derives Forward, Inverse, and Parameter tasks. Graded continuous proximity
  can reward a Solver that remains fundamentally incorrect, broadcasting
  distorted learnability signals back to the Proposer.}
  \label{fig:naive-self-play}
\end{figure}

However, naively mapping this paradigm to physical reasoning reveals a
fundamental structural breakdown (Figure~\ref{fig:naive-self-play}). When a
shared Proposer--Solver policy proposes rigid-body scenes and evaluates a single
simulated trajectory using continuous proximity scores, two failure modes
emerge (Section~\ref{sec:motivation}): (1) single-trajectory readouts cannot
construct or verify intervention tasks, starving the curriculum of causal
supervision; and (2) continuous proximity rewards inflate learnability signals
for uniformly incorrect rollouts, inducing reward hacking without improving
exact physical correctness. These findings establish two indispensable
requirements for physical self-play: the verification environment must
actively execute interventions across paired branches, and derived visual
tasks must be bound to discrete, exact ground truth.

We introduce \physevolve{}, a verifier-grounded self-evolving framework that
learns causal physical reasoning from policy-proposed executable worlds and
simulation-authored tasks (Figure~\ref{fig:physevolve-framework}). Rather than
authoring text questions directly, a task-conditioned Proposer generates an
executable \emph{world program} and an \emph{evidence schema}. A deterministic
physics executor validates and freezes the scene, executes matched baseline and
body-elided counterfactual branches under identical initial conditions, and
derives an answer-bound question from their divergent outcomes. The shared VLM
then solves the task from the baseline video, receiving exact discrete feedback.
This architecture establishes a strict information boundary: the policy guides
the exploration of physical environments, while the fixed simulator alone
determines physical validity and factual/counterfactual ground truth. The
resulting exact solver feedback simultaneously drives shared-policy updates via
GRPO and guides an adaptive curriculum over physical reasoning intents.

Training on executor-synthesized tasks over policy-proposed worlds produces
targeted transfer across three physical-reasoning benchmarks without any
human-annotated intervention data. Starting from Qwen3-VL-8B-Instruct
\citep{bai2025qwen3vl}, \physevolve{} improves PhysBench Throwing by $8.78$
percentage points and Judgment by $5.08$ points. Crucially, on ContPhy,
performance on Counterfactual Removal and Goal-Pit increases by $7.38$ and
$24.44$ points, transferring causal intervention principles zero-shot to
held-out continuous media (cloth, fluid, rope, and soft bodies) entirely absent
from the rigid-body training templates. On real-world videos in CausalPhys,
\physevolve{} likewise achieves consistent gains on Anticipation ($+2.46$) and
Intervention ($+2.27$).

Our core contributions are summarized as follows:
\begin{enumerate}[leftmargin=18pt,itemsep=1pt,parsep=0pt,topsep=2pt,partopsep=0pt]
  \item \textbf{Verifier-grounded physical self-play:} We present \physevolve{},
  the first self-evolving framework that trains VLMs for causal physical
  reasoning via model-proposed executable worlds, enforcing a strict information
  boundary between policy exploration and deterministic simulator verification.
  \item \textbf{Intervention-grounded task synthesis:} We introduce an
  execution pipeline that executes matched baseline and body-elided branches
  under identical seeds, transforming continuous multi-body dynamics into
  discrete, answer-bound multi-choice tasks with exact verifier rewards.
  \item \textbf{Out-of-distribution causal transfer \& audit:} We demonstrate
  that self-play on rigid-body programs transfers zero-shot to complex
  continuous dynamics and real-world video, and provide an empirical audit
  exposing why naive single-trajectory physical self-play fundamentally collapses.
\end{enumerate}


\section{Related Work}
\label{sec:related}

\paragraph{Physical reasoning in vision-language models.}
\looseness=-1
Extensive studies reveal that current multimodal models handle static visual
appearance and geometric arrangements far more reliably than temporal
prediction, physical intervention, and counterfactual reasoning
\citep{riochet2018intphys,bear2021physion,tung2023physionpp,yi2020clevrer,
ates2022craft,chow2025physbench,tang2026causalphys}. Broader multimodal
reasoning benchmarks further expose that vision-language models remain
unreliable on structured logical reasoning \citep{yuan2025mmereasoning},
motivating verifiable, grounded supervision rather than perception-only
training. To diagnose physical understanding beyond passive prediction,
counterfactual benchmarks introduce paired simulator-generated alternative
trajectories and intervention questions
\citep{baradel2020cophy,hong2022comphy,patel2022cripp}. Recently,
simulator-grounded training has emerged to convert simulation traces into
verifiable supervision: notably, Sim2Reason synthesizes physics QA pairs from
predefined simulator rollouts and filters shortcut solutions
\citep{prabhudesai2026sim2reason}. Unlike offline trajectory-to-QA pipelines,
\physevolve{} autonomously synthesizes verifiable counterfactual supervision
\emph{online} by executing paired intervention branches over active,
policy-proposed physical worlds.

\paragraph{Self-evolving agents and environment co-design.}
While standard self-training bootstraps reasoning over fixed problem sets
\citep{zelikman2022star,gulcehre2023rest}, self-evolving agents generate the
very curricula and environments from which they learn. Absolute Zero pioneered
shared-policy proposer--solver self-play with code interpreters
\citep{zhao2025azr}, inspiring subsequent frameworks to expand task substrates
across multi-agent debates, geometric spaces, games, and tool-augmented
execution \citep{huang2025rzero,chen2025multiagentevolve,xia2025agent0,
liu2025spiral,wei2025selfplayswerl,he2025visplay,wang2025visionzero,
li2026spatialevo}. A complementary line of modular and agentic multimodal
systems---spanning visual-to-code generation, agentic coding for games,
search-augmented image generation, multimodal deep search, agentic perception,
and failure-aware software repair---demonstrates that decomposing perception,
planning, and execution into verifiable sub-routines scales reliably across
complex tool-use environments
\citep{jiang2026screencoder,jiang2026opengame,feng2026gensearcher,
jiao2026searcheyes,fan2026evirover}. In parallel, unsupervised environment design (UED) co-evolves
adaptive environments alongside control policies based on learner regret or
progress \citep{wang2019poet,dennis2020paired,parkerholder2022accel,zala2024envgen,
liang2024eurekaverse}. While existing self-play frameworks operate primarily in
discrete symbolic spaces, \physevolve{} shifts the frontier of self-evolution
to continuous multi-body physics, delegating the physical environment proposal to
the policy while retaining deterministic verification and task synthesis within
the executor.

\paragraph{Reinforcement learning with verifiable rewards.}
\looseness=-1
Reinforcement learning with verifiable rewards (RLVR) substitutes brittle,
exploitable learned reward models with deterministic, ground-truth verifiers
in domains admitting objective computation \citep{deepseek2025r1,lambert2025tulu3}.
Verification methods have expanded from exact answer matching and test-suite
execution to hybrid dense-sparse rewards, self-consistency checks, and formal
proof checking \citep{tao2026hero,zhao2026intuitor,wilf2025psv}. Recent
multimodal RL work further studies reward design under heterogeneous or
competing objectives, including multi-task visual reasoning with moving-average
reward normalization, thinking-with-image video reward models, and reasoning
reward models for agents
\citep{feng2025onethinker,wang2025vrthinker,fan2026evirover}. While RLVR has
achieved remarkable success in mathematical deduction and program synthesis,
extending objective verification to multimodal physical reasoning has remained
an open challenge. \physevolve{} realizes RLVR for multimodal physics by
grounding policy updates in discrete, simulation-verified causal outcomes rather
than heuristic continuous proximity metrics.


\section{Methodology}
\label{sec:method}

\looseness=-1
We introduce \physevolve{}, a verifier-grounded self-evolving framework that
learns causal physical reasoning by having a shared vision--language policy
propose executable worlds while a deterministic physics executor authors and
verifies derived multimodal tasks.
Below, Section~\ref{sec:motivation} first diagnoses the structural pitfalls of
naively transferring code self-play to physical reasoning and establishes two
necessary design principles.
Sections~\ref{sec:overview}--\ref{sec:training} then detail the framework
architecture, the task-conditioned proposer, the intervention-grounded task
synthesis pipeline, and the hierarchical co-evolution training loop.

\subsection{Problem Formulation and Pitfalls of Naive Physical Self-Play}
\label{sec:motivation}

\looseness=-1
In discrete symbolic domains such as mathematics and code synthesis,
self-play enables models to surpass fixed data limits by autonomously
generating and solving their own problems \citep{zhao2025azr}. The policy
evolves by optimizing a learnability reward that dynamically prioritizes tasks
near the boundary of its current competence. The viability of this paradigm
stems from its objectivity: a deterministic interpreter executes model-written
specifications and returns exact correctness signals without human supervision.

Given that grounded, intervention-level annotations are exceptionally scarce in
physical reasoning, transferring this self-play paradigm to multimodal physics
by substituting a physics simulator for the symbolic interpreter appears to be
a natural solution. In a direct physical transfer, a shared Proposer--Solver
policy interacts with a rigid-body simulator. For each proposed world
$\mathcal{W}$, the executor simulates one factual trajectory $\tau$ under seed
$\xi$, extracts an observation $x$ alongside a simulator-derived target $y$,
and scores the Solver's prediction $\hat y$ via a graded similarity metric $s$:
\begin{equation}
  \label{eq:naive}
  \tau=\Phi_\xi(\mathcal{W}),\qquad
  (x,y)=\operatorname{readout}(\tau),\qquad
  r_{\mathrm{naive}}=s(\hat y,y).
\end{equation}
Here $\Phi_\xi$ denotes deterministic simulation and $s$ assigns partial
credit based on continuous numerical or spatial proximity, intended to prevent
sparse, uninformative all-zero rewards early in training. While this transfer
ostensibly preserves both autonomous task generation and verifiable
supervision, empirical auditing reveals two fatal structural failure modes.

\textbf{{Failure Mode 1: Single-trajectory execution cannot ground causal interventions.}}
Physical reasoning distinctively demands understanding how scenes respond to
counterfactual interventions---outcomes that never occurred in the observed
trajectory \citep{pearl2009causality}. However, evaluating a single trajectory
$\tau$ restricts the task space strictly to passive observational readouts
(e.g., predicting an object's future position or initial velocity). In our audit
of the naive transfer baseline, exactly zero intervention or counterfactual
tasks were synthesized across 100 training rounds, causing PhysBench dynamics
accuracy to stagnate and reasoning performance to degrade
(Section~\ref{sec:naive-diagnosis}). This limitation is structural: causal
ground truth cannot be inferred from a single trajectory without paired,
counterfactual branch executions. Neither scaling model capacity nor expanding
rollout volume can produce verified labels that lie beyond the verification
interface itself.

\textbf{{Failure Mode 2: Graded proximity rewards induce learnability illusions.}}
In code and math, execution verification yields a rigorous binary correctness
signal. In continuous physics, however, substituting exact match with graded
proximity $s$ fundamentally distorts reinforcement signals. Because $s$ scores
relative closeness rather than true task resolution, clusters of uniformly
incorrect rollouts (e.g., predicting stationary velocities or median parameters)
yield intermediate rewards. These reward distributions artificially inflate the
empirical learnability metric $r_{\mathrm{learn}} = 1 - 2|\bar a - 0.5|$,
misleading the curriculum into classifying failed, trivial tasks as productive
learning boundaries. Consequently, graded reward steadily increases during naive
training while exact physical correctness remains near zero
(Section~\ref{sec:naive-diagnosis}, Table~\ref{tab:naive-diagnosis}).

\looseness=-1
These failure modes establish that continuous physics cannot be treated as a
trivial drop-in replacement for discrete code interpreters. Autonomous physical
reasoning requires two non-negotiable architectural guarantees:
\begin{enumerate}[leftmargin=18pt,itemsep=1pt,parsep=0pt,topsep=2pt,partopsep=0pt]
  \item \textbf{Branch-level verification:} The executor must actively execute
  paired, counterfactual interventions (divergent simulation branches) under
  identical initial conditions to establish objective causal ground truth.
  \item \looseness=-1 \textbf{Discrete answer binding:} Derived physical tasks must map
  continuous trajectories into discrete, answer-bound problems, ensuring that
  Solver success reflects exact correctness rather than deceptive proximity.
\end{enumerate}

\subsection{Framework Overview}
\label{sec:overview}

\begin{figure}[!tbp]
  \centering
  \setlength{\abovecaptionskip}{5pt}
  \includegraphics[width=0.95\linewidth,trim=0 12 0 16,clip]{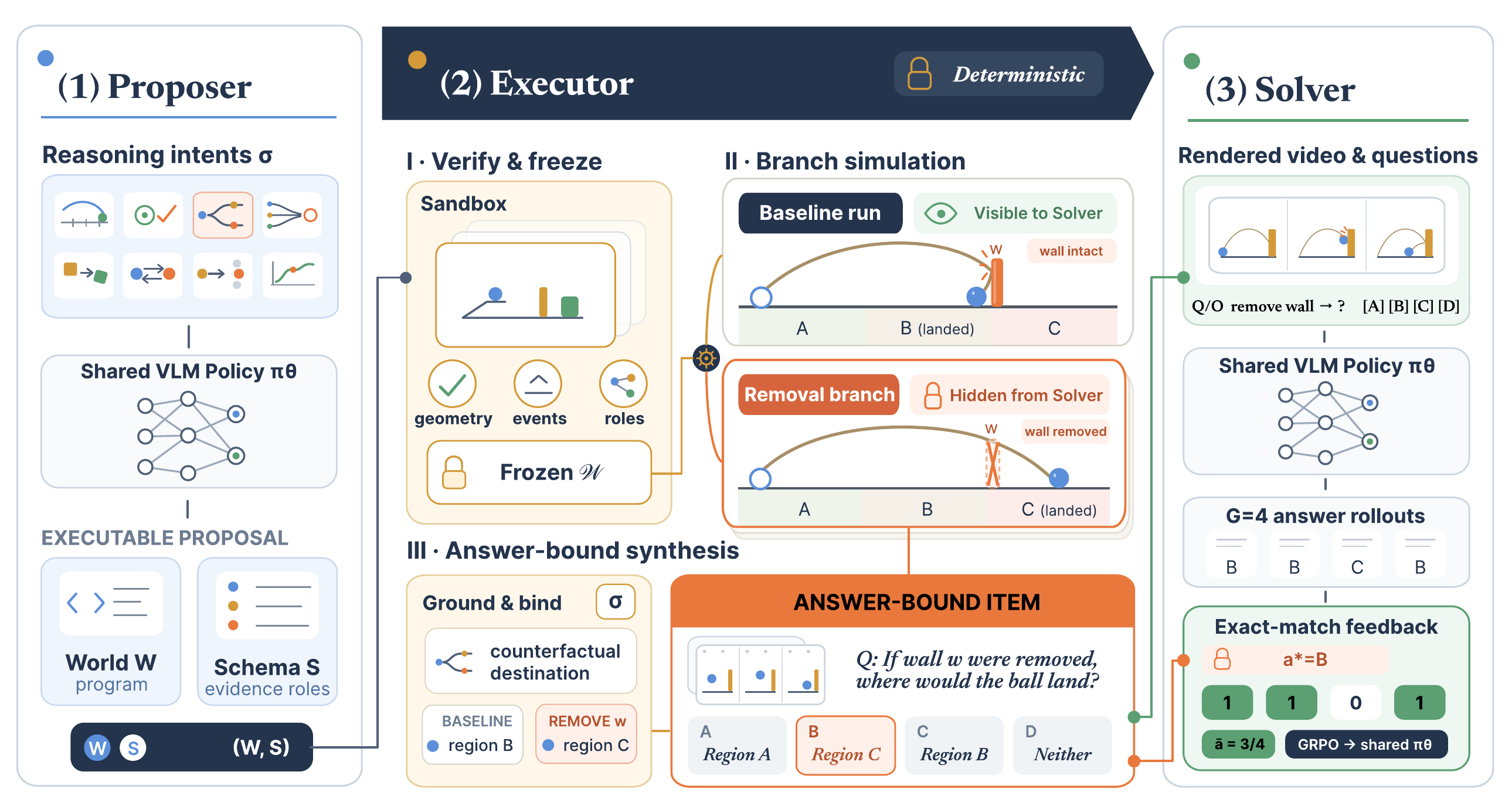}
  \caption{Overview of \physevolve{}. A shared VLM proposes an executable
  world program and evidence schema. A fixed deterministic executor validates
  and freezes the scene, simulates matched baseline and body-elided branches
  when required, and synthesizes an answer-bound item. Exact discrete Solver
  feedback updates the shared policy and the curriculum.}
  \label{fig:physevolve-framework}
\end{figure}

Guided by these requirements, we introduce \physevolve{}
(Figure~\ref{fig:physevolve-framework}). At each round, a
\emph{Task-Conditioned Proposer} transforms a sampled reasoning intent into an
executable candidate world. Through \emph{Intervention-Grounded Task Synthesis},
a fixed executor establishes factual and counterfactual outcomes via
deterministic simulation and synthesizes discrete, answer-bound multiple-choice
questions. Finally, \emph{Co-Evolution Training} translates exact Solver
correctness into joint updates for the shared policy and an adaptive curriculum.

Let $\pi_\theta$ denote the shared vision--language policy. The identical
parameters serve as Proposer $\pi_\theta^{\mathrm{P}}$ and Solver
$\pi_\theta^{\mathrm{S}}$. The Proposer controls the exploration of executable
physical environments, while the Solver receives only rendered baseline video
observations and the resulting question and options. Crucially, a strict
\emph{information boundary} is enforced: the fixed physics executor alone
accesses internal simulator states, computes counterfactual branches, and
binds the correct answer. This boundary enables the policy to actively shape
its own training distribution without ever setting its own training labels,
eliminating label contamination and reward hacking.

\subsection{Task-Conditioned Proposer}
\label{sec:proposer}

The Proposer receives a reasoning intent $\sigma$, which specifies the target
physical relation to probe and the evidence required to make that relation
uniquely decidable. Grounded in classical cognitive physics
\citep{battaglia2013simulation,spelke1990principles}, we instantiate
simulation across three fundamental Newtonian paradigms: (1) projectile motion
and parabolic dynamics (\emph{Throwing}), (2) impulse-momentum transfer and
kinematic collisions (\emph{Collision}), and (3) continuous kinematics and
contact interactions (\emph{Motion}), spanning eight active reasoning intents.

\looseness=-1
Conditioned on $\sigma$, the Proposer samples a tuple
$(\mathcal{W},\mathcal{S})\sim\pi_\theta^{\mathrm{P}}(\cdot\mid\sigma)$:
an executable \emph{world program} $\mathcal{W}$ and an \emph{evidence schema}
$\mathcal{S}$. The world program is authored in PhysDSL (a domain-specific
Python API over rigid-body simulation) specifying 3D geometry, material
properties (mass, restitution, friction), initial states, kinematic
constraints, and rendering viewpoints. The evidence schema $\mathcal{S}$
declares the semantic roles of key entities (e.g., query projectiles,
barrier obstacles, potential collision targets) required by intent $\sigma$.
The executor validates these role declarations against the realized simulation,
rejecting any proposal whose physical dynamics cannot support the requested
reasoning intent. Complete specifications of PhysDSL and evidence schemas are
provided in Appendices~\ref{app:dsl} and~\ref{app:spec}.

\subsection{Intervention-Grounded Task Synthesis}
\label{sec:executor}

A candidate pair $(\mathcal{W},\mathcal{S})$ represents only a potential
reasoning opportunity; simulation execution must establish its physical
semantics before it can constitute a valid training task. The executor admits
an item only when execution recovers unambiguous physical evidence that
determines a unique answer. This contract is executed across three sequential
stages:

\textbf{World verification and freezing.}
The executor parses and executes the world program in a sandboxed simulator,
resolving evidence-schema roles against simulated rigid bodies. It admits
only worlds exhibiting valid geometry, non-trivial physical events, and the
causal prerequisites of intent $\sigma$. A bounded deterministic repair stage
corrects minor placement or boundary syntax violations without introducing new
bodies or joints. Once admitted, the world definition is strictly \emph{frozen}
into an immutable scene definition. Freezing guarantees that all subsequent
intervention branches share identical initial geometry, material properties,
and simulation seeds, ensuring that divergent outcomes are strictly attributable
to declared interventions.

\textbf{Branch elision.}
For interventional reasoning intents, the executor constructs a finite set of
body removals $\mathcal{B}_\sigma$. For each $B\in\mathcal{B}_\sigma$, it elides
those bodies and their dependent constraints from the frozen world, yielding
the matched execution pair:
\begin{equation}
  \label{eq:branch}
  \bigl(\tau_{\emptyset},\tau_{B}\bigr)
  =
  \bigl(\Phi_{\xi}(\mathcal{W}),
        \Phi_{\xi}(\mathcal{W}\setminus B)\bigr),
  \qquad B\in\mathcal{B}_\sigma.
\end{equation}
Both trajectories share the identical simulator state, random seed $\xi$, and
time horizon. The Solver observes only a video rendering of the factual
trajectory $\tau_{\emptyset}$, while the executor retains the divergent branch
trajectories $\{\tau_B\}$ as objective ground truth.

\textbf{Answer-bound question synthesis.}
Each reasoning intent defines a discrete readout function, feasibility margins,
and natural language templates. The executor applies the readout across baseline
and branch trajectories. An item is suppressed if readouts are non-unique,
infeasible, or fall within sensitive physical margin thresholds (e.g., velocity
differences below $0.30$\,m/s). Otherwise, the executor determines the unique
factual or counterfactual outcome, binds it to ground-truth option index
$a^\star\in\{A,B,C,D\}$, and renders the final visual question and distractors.
For inverse intervention tasks (e.g., ``Which wall should be removed to make
the ball land in the target pit?''), the executor runs all candidate elisions
(remove A, remove B, remove both) and validates whether exactly one branch
achieves the goal; ``none of the above'' is selected only if no branch reaches
the target. Detailed contracts, margin tests, and binding rules are documented
in Appendix~\ref{app:families}.

\subsection{Co-Evolution Training and Adaptive Curriculum}
\label{sec:training}

\looseness=-1
The synthesized answer-bound items close the self-play loop by providing exact,
discrete supervision to both policy roles.

\textbf{Exact solver supervision.}
\looseness=-1
Each retained item defines a group of $G$ Solver rollouts from policy
$\pi_\theta^{\mathrm{S}}$. Given the executor-bound label $a^\star$ and predicted
options $\{\hat a_i\}_{i=1}^G$, the binary Solver reward and empirical item
success rate are:
\begin{equation}
  \label{eq:selfplay-reward}
  r_i^{\mathrm{solve}} = \mathbf{1}[\hat a_i=a^\star],
  \qquad
  \bar a = \frac{1}{G}\sum_{i=1}^{G} r_i^{\mathrm{solve}}.
\end{equation}
In contrast to the continuous proximity score $s$ in Equation~\eqref{eq:naive},
$r_i^{\mathrm{solve}}$ provides unambiguous binary verification immune to
proximity-induced reward hacking.

\textbf{Hierarchical co-evolution.}
\looseness=-1
Co-evolution coordinates task selection and world exploration across two
complementary levels:
\begin{itemize}[leftmargin=18pt,itemsep=1pt,parsep=0pt,topsep=2pt,partopsep=0pt]
  \item \emph{Macro-level intent curriculum:} Across reasoning intents, the
  rolling Solver accuracy $\bar a_f$ for each task family $f$ governs adaptive
  intent sampling. After an 8-round cold start, intent sampling weights follow
  $w_f = \rho_f \max(0.05, 4\bar a_f(1 - \bar a_f))$, where $\rho_f$ denotes
  structural family priors. This formulation concentrates training on intents
  situated at the frontier of the model's evolving competence (where
  $4\bar a_f(1-\bar a_f)$ peaks near $\bar a_f=0.5$). A family-difficulty
  multiplier further modulates Proposer credit according to rolling intent
  challenge.
  \item \emph{Micro-level world exploration:} Conditioned on the sampled intent
  $\sigma$, the Proposer explores the space of valid, physically rich world
  programs. Proposer rewards reward successful compilation, dynamic collision
  feasibility, and structural edit-distance diversity $I$, penalized by
  anti-copy and saturation controls (Appendix~\ref{app:reward}).
\end{itemize}

\textbf{Shared-policy GRPO optimization.}
\looseness=-1
Both Proposer and Solver roles are optimized jointly via Group Relative Policy
Optimization (GRPO) on the shared parameters $\pi_\theta$, regularized against
a frozen reference policy $\pi_{\mathrm{ref}}$. Proposer groups share a sampled
intent $\sigma$, while Solver groups share an executor-rendered item. Joint
optimization ensures that while the Solver masters causal physical deduction,
the Proposer continuously adapts to author executable environments that remain
informative and verifiable (Appendix~\ref{app:hparams}).

\section{Experiments}
\label{sec:experiments}

\subsection{Experimental Setup}
\label{sec:experimental-setup}

\paragraph{Research questions.}
\looseness=-1
We design our empirical evaluation around four central questions:
\textbf{(Q1) Targeted capability transfer:} Does \physevolve{} yield targeted
improvements on the specific physical capabilities it simulates during training?
\textbf{(Q2) Out-of-distribution transfer:} Do causal principles learned in
synthetic rigid-body environments generalize to unseen complex media (fluid,
cloth, soft bodies) and real-world video?
\textbf{(Q3) Architectural necessity:} How critically do paired branch
simulation and joint Proposer--Solver optimization contribute to performance?
\textbf{(Q4) Failure anatomy:} What exact supervisory signals are forfeited by
relying on single-trajectory execution and continuous proximity rewards?

\paragraph{Training configuration.}
All experimental variants are initialized from the open-weight base model
Qwen3-VL-8B-Instruct \citep{bai2025qwen3vl}. Training is executed in a single
run for 100 steps using one-epoch GRPO with a learning rate of $10^{-6}$. At
each step, the curriculum samples $N=4$ reasoning intents and generates $G=4$
world program proposals per intent, capping retention at 32 active Solver groups
with 4 rollouts per group. The complete hyperparameters and training loop are
detailed in Appendix~\ref{app:hparams} and Algorithm~\ref{alg:loop}.

\paragraph{Benchmarks and evaluation protocols.}
We evaluate across three rigorous physical reasoning benchmarks:
(1) \textbf{PhysBench} \citep{chow2025physbench} covers broad static and
dynamic physical QA over 9,802 items;
(2) \textbf{ContPhy} \citep{zheng2024contphy} benchmarks counterfactual and
goal-directed physical reasoning across 1,950 scenarios involving complex
continuous dynamics (cloth, fluid, rope, and soft bodies); and
(3) \textbf{CausalPhys} \citep{tang2026causalphys} assesses physical anticipation
and intervention in 3,057 real-world video scenarios.
All evaluated models share identical prompts, visual frame sampling, greedy
decoding, and official evaluation protocols (Appendix~\ref{app:evaluation-protocols}).

\paragraph{Baselines and comparative context.}
Our primary, strictly matched comparison comprises:
(1) the base \textbf{Qwen3-VL-8B-Instruct} model;
(2) \textbf{Naive self-play}, which replicates single-trajectory execution with
continuous graded proximity rewards; and
(3) \textbf{\physevolve{}}.
To contextualize performance across model scales, we also report unaligned
results for prominent open-weight models: InternVL2.5-8B \citep{chen2024internvl25},
Llama-3.2-11B-Vision \citep{meta2024llama32vision}, Pixtral-12B \citep{agrawal2024pixtral},
Qwen2.5-VL-7B \citep{bai2025qwen25vl}, and the scaled Qwen3-VL-32B \citep{bai2025qwen3vl}.

\subsection{Main Results: Targeted and External Transfer}
\label{sec:main-results}

Table~\ref{tab:main-results} presents overall and capability-specific accuracies
across the three benchmarks, separating matched 8B configurations from
open-weight context models.

\begin{table}[!t]
  \centering
  \setlength{\belowcaptionskip}{5pt}
  \caption{\looseness=-1 Accuracy (\%) across the three physical reasoning benchmarks. The upper
  block provides contextual reference models; bold text indicates the best
  performance within the strictly matched 8B comparison block.}
  \label{tab:main-results}
  \begingroup
  \scriptsize
  \setlength{\tabcolsep}{1.0pt}
  \renewcommand{\arraystretch}{1.0}
  \resizebox{\linewidth}{!}{%
  \begin{tabular}{l@{\hspace{3pt}}*{4}{c}@{\hspace{5pt}}*{3}{c}@{\hspace{5pt}}*{3}{c}}
    \toprule
    & \multicolumn{4}{c}{PhysBench}
    & \multicolumn{3}{c}{CausalPhys}
    & \multicolumn{3}{c}{ContPhy} \\
    \cmidrule(lr){2-5}\cmidrule(lr){6-8}\cmidrule(l){9-11}
    Model
    & Ovr. & Throw. & Motion & Judg.
    & Ovr. & Antic. & Interv.
    & Ovr. & CF-Rm. & Goal-Pit \\
    \midrule
    \multicolumn{11}{c}{\emph{Contextual open-weight references}} \\
    InternVL2.5-8B~\citep{chen2024internvl25}
    & 43.8 & 40.0 & 27.7 & 32.7
    & 50.2 & 51.5 & 42.4
    & 31.4 & 48.4 & 23.3 \\
    Llama-3.2-11B-Vision~\citep{meta2024llama32vision}
    & 38.7 & 36.6 & 27.2 & 30.8
    & 44.2 & 44.1 & 40.1
    & 20.7 & 0.0 & 20.0 \\
    Pixtral-12B~\citep{agrawal2024pixtral}
    & 46.9 & 41.2 & 35.2 & 31.0
    & 50.2 & 53.0 & 44.5
    & 21.5 & 40.2 & 13.3 \\
    Qwen2.5-VL-7B~\citep{bai2025qwen25vl}
    & 49.4 & 44.4 & 50.6 & 33.1
    & 52.6 & 56.5 & 45.7
    & 25.9 & 11.5 & 24.4 \\
    Qwen3-VL-32B~\citep{bai2025qwen3vl}
    & 60.9 & 47.1 & 83.5 & 44.3
    & 58.9 & 61.9 & 52.9
    & 41.0 & 45.9 & 26.7 \\
    \midrule
    \multicolumn{11}{c}{\emph{Matched Qwen3-VL-8B comparison}} \\
    Qwen3-VL-8B (base)~\citep{bai2025qwen3vl}
    & 52.9 & 46.1 & 54.7 & 34.9
    & 53.9 & 55.9 & 42.8
    & 37.1 & 44.3 & 20.0 \\
    Naive self-play
    & 52.4 & 45.9 & 55.5 & 35.2
    & \textbf{55.8} & 57.6 & \textbf{45.9}
    & 37.4 & 43.4 & 25.6 \\
    \physevolve{} (ours)
    & \textbf{53.5}\,\posd{0.6}
    & \textbf{54.9}\,\posd{8.8}
    & \textbf{56.5}\,\posd{1.7}
    & \textbf{40.0}\,\posd{5.1}
    & 54.5\,\posd{0.6}
    & \textbf{58.4}\,\posd{2.5}
    & 45.0\,\posd{2.3}
    & \textbf{39.5}\,\posd{2.5}
    & \textbf{51.6}\,\posd{7.4}
    & \textbf{44.4}\,\posd{24.4} \\
    \bottomrule
  \end{tabular}%
  }
  \endgroup
  \par\vspace{6pt}
  \begin{minipage}{\linewidth}
  \makeatletter\def\@captype{figure}\makeatother
  \centering
  \setlength{\abovecaptionskip}{5pt}
  \includegraphics[width=0.83\linewidth,trim=0 3 0 5,clip]{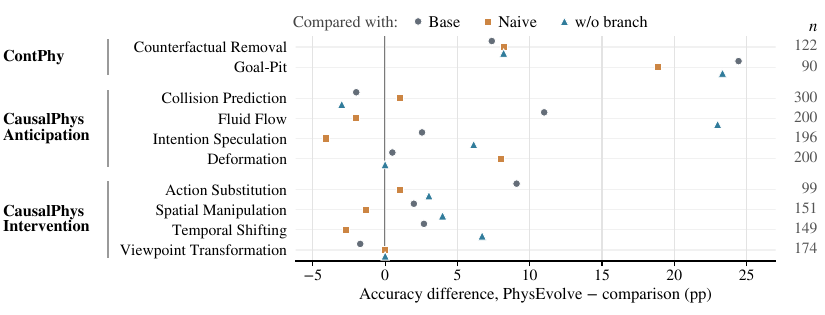}
  \caption{Task-level transfer across matched comparisons. Markers show accuracy
  differences between \physevolve{} and each comparison model on ContPhy
  (Counterfactual Removal, Goal-Pit) and CausalPhys subcategories. Positive
  values favor \physevolve{}. Sample counts appear at right; see
  Table~\ref{tab:causalphys-fine-grained} for complete accuracies.}
  \label{fig:transfer-task-differences}
\end{minipage}
\end{table}

\paragraph{Q1: Targeted capability transfer on PhysBench.}
Within the matched 8B evaluation, \physevolve{} achieves substantial gains on
tasks directly aligned with simulated Newtonian paradigms, improving Throwing by
$8.78$ percentage points, Judgment by $5.08$, and Motion by $1.74$. In stark
contrast, Naive self-play yields negligible or negative differences on these
categories ($-0.24$ on Throwing, $+0.26$ on Judgment). Crucially, \physevolve{}
improves overall accuracy ($+0.55$) without catastrophic forgetting of general
visual capabilities: performance remains close to or above the base model across most non-target partitions, including static relations and object identification (Appendix~\ref{app:physbench-breakdown},
Table~\ref{tab:physbench-complete-taxonomy}).

\paragraph{Q2: Out-of-distribution transfer to complex media and real video.}
\looseness=-1
The most compelling evidence of physical generalization emerges on ContPhy:
despite being trained exclusively on synthetic rigid-body simulations,
\physevolve{} demonstrates remarkable zero-shot transfer to continuous media
(cloth, fluid, rope, and soft bodies). On Counterfactual Removal, accuracy
increases by $7.38$ points over Base ($51.64\%$ vs.\ $44.26\%$), while on
Goal-Pit, accuracy surges by $24.44$ points ($44.44\%$ vs.\ $20.00\%$,
surpassing even Qwen3-VL-32B's $26.7\%$). On CausalPhys real-world videos,
\physevolve{} similarly yields consistent gains on Anticipation ($+2.46$) and
Intervention ($+2.27$). Because the non-rigid dynamics and visual textures of
ContPhy are completely absent from the training templates, these gains confirm
that \physevolve{} internalizes an abstract \emph{intervention operator}---the
ability to deduce how physical states diverge under structural removal---rather
than memorizing superficial simulation artifacts.

\begin{figure}[!tbp]
  \centering
  \setlength{\abovecaptionskip}{5pt}
  \includegraphics[width=0.83\linewidth,trim=0 2 0 1,clip]{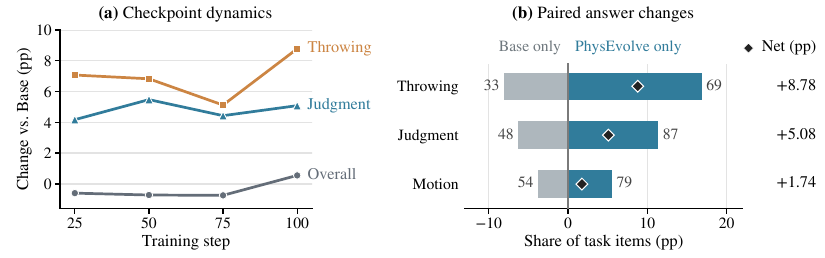}
  \caption{PhysBench training dynamics and paired correctness analysis.
  Left: accuracy gains over Base across scheduled training checkpoints.
  Right: item-level paired analysis at the terminal checkpoint, separating
  questions answered correctly only by \physevolve{} from those answered
  correctly only by Base. Diamonds indicate net correct item gains; see
  Table~\ref{tab:physbench-paired} for complete counts.}
  \label{fig:checkpoint-dynamics}
\end{figure}

\paragraph{Fine-grained task transfer and checkpoint dynamics.}
Figure~\ref{fig:transfer-task-differences} breaks down transfer across fine-grained
causal subcategories. Relative to the Base model, \physevolve{} improves Fluid Flow
by $11.00$ points and Action Substitution by $9.09$ points, significantly
outperforming the branch-disabled control (by $23.00$ points on Fluid Flow).
Furthermore, longitudinal checkpoint analysis (Figure~\ref{fig:checkpoint-dynamics}, left)
shows consistent gains over Base on Throwing and Judgment throughout self-play. At the
terminal checkpoint (Figure~\ref{fig:checkpoint-dynamics}, right), item-level
paired auditing reveals that \physevolve{} correctly answers 69 Throwing items that
Base misses against only 33 reversals (a net gain of 36 items), and net gains of 39
and 25 items on Judgment and Motion, confirming genuine competence acquisition.

\subsection{Ablation Studies}
\label{sec:ablations}

\paragraph{Q3: Necessity of branch-enabled synthesis and joint co-evolution.}
We investigate the architectural components of \physevolve{} via two targeted
endpoint ablations:

\begin{table}[!tbp]
  \centering
  \setlength{\belowcaptionskip}{5pt}
  \caption{\looseness=-1 Ablation of the complete branch-enabled training path. Evaluating
  endpoint accuracy (\%) under identical protocols when branch execution and
  dependent Counterfactual/Inverse families are deactivated.}
  \label{tab:ablations}
  \scriptsize
  \setlength{\tabcolsep}{2.5pt}
  \renewcommand{\arraystretch}{1.0}
  \begin{tabular*}{\linewidth}{@{\extracolsep{\fill}}lccrrccccc@{}}
    \toprule
    & \multicolumn{2}{c}{Training Path}
    & \multicolumn{2}{c}{Realized Exposure}
    & \multicolumn{2}{c}{CausalPhys}
    & \multicolumn{3}{c}{ContPhy} \\
    \cmidrule(lr){2-3}\cmidrule(lr){4-5}
    \cmidrule(lr){6-7}\cmidrule(l){8-10}
    Configuration
    & Branch Sim. & CF/Inv
    & Groups & Rollouts
    & Ovr. & Interv.
    & Ovr. & CF-Rm. & Goal-Pit \\
    \midrule
    \physevolve{}
    & On & On
    & 2,167 & 8,668
    & \textbf{54.50} & \textbf{45.03}
    & \textbf{39.54} & \textbf{51.64} & \textbf{44.44} \\
    \quad w/o branch-enabled path
    & Off & Off
    & 1,990 & 7,960
    & 51.42 & 41.71
    & 37.28 & 43.44 & 21.11 \\
    \midrule
    Difference
    & -- & --
    & +177 & +708
    & \textbf{+3.07} & \textbf{+3.32}
    & \textbf{+2.26} & \textbf{+8.20} & \textbf{+23.33} \\
    \bottomrule
  \end{tabular*}
\end{table}

\begin{table}[!tbp]
  \centering
  \setlength{\abovecaptionskip}{0pt}
  \setlength{\belowcaptionskip}{0pt}
  \sbox{\jointcaptionbox}{\begin{minipage}[t]{0.66\linewidth}
  \caption{Ablation of joint Proposer--Solver optimization. Disabling Proposer
  parameter updates while updating only Solver parameters collapses policy
  distribution and answer parse rates (\%).}
  \label{tab:joint-role-updates}
  \end{minipage}}
  \sbox{\auditcaptionbox}{\begin{minipage}[t]{0.32\linewidth}
  \caption{Training-time audit of the single-trajectory naive verifier across 100 rounds.}
  \label{tab:naive-diagnosis}
  \end{minipage}}
  \setlength{\pairedcaptionheight}{\dimexpr\ht\jointcaptionbox+\dp\jointcaptionbox\relax}
  \ifdim\dimexpr\ht\auditcaptionbox+\dp\auditcaptionbox\relax>\pairedcaptionheight
    \setlength{\pairedcaptionheight}{\dimexpr\ht\auditcaptionbox+\dp\auditcaptionbox\relax}
  \fi
  \parbox[t][\pairedcaptionheight][t]{0.66\linewidth}{\usebox{\jointcaptionbox}}\hfill
  \parbox[t][\pairedcaptionheight][t]{0.32\linewidth}{\usebox{\auditcaptionbox}}
  \par\vspace{6pt}
  \begin{minipage}[t]{0.66\linewidth}
  \vspace{0pt}
  \scriptsize
  \setlength{\tabcolsep}{1.5pt}
  \renewcommand{\arraystretch}{1.15}
  \begin{tabular*}{\linewidth}{@{\extracolsep{\fill}}lcccccc@{}}
    \toprule
    & \multicolumn{2}{c}{Training updates}
    & \multicolumn{2}{c}{CausalPhys}
    & \multicolumn{2}{c}{ContPhy} \\
    \cmidrule(lr){2-3}\cmidrule(lr){4-5}\cmidrule(l){6-7}
    Configuration & Proposer & Solver & ACC & Parse rate & ACC & Parse rate \\
    \midrule
    \physevolve{} & $\checkmark$ & $\checkmark$
      & \textbf{54.50} & \textbf{97.84} & \textbf{39.54} & \textbf{100.00} \\
    w/o Proposer-role updates & -- & $\checkmark$
      & 15.01 & 26.07 & 17.85 & 99.90 \\
    \midrule
    $\Delta$ & -- & -- & $+39.48$ & $+71.77$ & $+21.69$ & $+0.10$ \\
    \bottomrule
  \end{tabular*}
  \end{minipage}\hfill
  \begin{minipage}[t]{0.32\linewidth}
  \vspace{0pt}
  \scriptsize
  \setlength{\tabcolsep}{1pt}
  \renewcommand{\arraystretch}{1.15}
  \begin{tabular*}{\linewidth}{@{\extracolsep{\fill}}lccc@{}}
    \toprule
    \shortstack[l]{Prediction\\task} & \shortstack{Mean graded\\reward}
      & \shortstack{Exact\\accuracy} & \shortstack{Zero-exact\\rounds} \\
    \midrule
    Forward & 0.373 & 0.00\% & 100/100 \\
    Inverse & 0.664 & 6.63\% & 75/100 \\
    Parameter & 0.787 & 5.44\% & 83/100 \\
    \bottomrule
  \end{tabular*}
  \end{minipage}
\end{table}

\textbf{Essentiality of paired branch execution.}
In Table~\ref{tab:ablations}, removing branch simulation and its dependent
Counterfactual/Inverse task families sharply degrades downstream causal transfer.
The performance drop is especially drastic on ContPhy Counterfactual Removal
($-8.20$ points) and Goal-Pit ($-23.33$ points), dropping performance back to the
un-finetuned Base level. This ablation empirically confirms that paired branch
elision is the indispensable engine for acquiring causal physical reasoning.

\textbf{Criticality of joint Proposer--Solver optimization.}
In Table~\ref{tab:joint-role-updates}, we freeze Proposer role updates, training
the shared policy exclusively via Solver rollouts. Without co-evolution, the
policy undergoes severe representational collapse: accuracy drops precipitously
to $15.01\%$ on CausalPhys and $17.85\%$ on ContPhy, accompanied by a collapse in
answer parse rate (falling to $26.07\%$ on CausalPhys and $0\%$ in later steps).
Updating the Proposer keeps world generation diverse and physically non-trivial,
preventing the Solver from over-adapting to a degenerate data stream.

\subsection{Auditing the Single-Trajectory Interface}
\label{sec:naive-diagnosis}

\paragraph{Q4: What causes naive physical self-play to collapse?}
\looseness=-1
We conduct an empirical audit of the single-trajectory verifier and task stream
in Naive self-play to validate the theoretical pitfalls identified in
Section~\ref{sec:motivation}.

\textbf{Divergence between graded reward and exact correctness.}
Table~\ref{tab:naive-diagnosis} exposes a striking pathology: continuous proximity
rewards $s$ consistently provide deceptive reinforcement. For Forward prediction,
rollouts achieve a high mean graded reward of $0.373$ despite \emph{zero} exact
matches across all 100 rounds. For Inverse and Parameter tasks, mean rewards reach
$0.664$ and $0.787$ while exact accuracy remains below $7\%$. An error audit
reveals severe mode collapse: the policy learns to output stationary initial
velocities (746 instances) or the median rung of the parameter ladder (145 instances),
exploiting the graded metric to reap partial credit without solving the physical
events.

\textbf{Task space starvation.}
Over 100 training rounds, the naive single-trajectory verifier generated
\emph{zero} intervention questions and \emph{zero} counterfactual tasks.
Constrained to an un-intervened factual trajectory $\tau$, the system is incapable
of synthesizing causal supervision. Consequently, downstream PhysBench dynamics
accuracy stagnates (Table~\ref{tab:main-results}). 

\section{Conclusion}
\label{sec:discussion}


\physevolve{} demonstrates that verifier-grounded self-evolution extends to continuous
multimodal physical reasoning by anchoring self-play in executable 3D world programs
and paired counterfactual branches. Converting continuous physics into discrete verifier
contracts eliminates the dual traps of continuous reward hacking and causal blindness
inherent in naive physical self-play. Strong targeted gains on Newtonian dynamics and zero-shot transfer to unseen continuous media and real-world video show that counterfactual execution provides a
scalable foundation for generalizable physical intelligence in vision-language models, entirely without any human-annotated physical reasoning supervision during self-play training. Each training target is bound to a verified physical outcome.

\microtypesetup{protrusion=false,expansion=false}
\bibliography{iclr2027_conference}
\bibliographystyle{plainnat}

\newpage
\appendix

\section{Complete PhysDSL and Executable World Interface}
\label{app:dsl}

\subsection{Complete callable surface}

A proposal is a Python module containing \texttt{build\_scene(api)} and a
literal \texttt{SCENE\_MANIFEST}. Every construction/setter call mutates a
builder and returns \texttt{None}; the proposal cannot branch on simulator
state. The executor alone calls \texttt{freeze()} to obtain a scene. Table
\ref{tab:physdsl-api} lists the complete public surface used by the current
  executor, including operations not exercised by every reported world class.
  Section~\ref{app:gallery} instantiates the interface with three auditable
  golden-world illustrations.

\begin{table}[h]
  \centering
  \caption{Complete PhysDSL callable surface. Arguments are validated when the
  call is made; \texttt{freeze()} returns the immutable executor input.}
  \label{tab:physdsl-api}
  \small
  \setlength{\tabcolsep}{5pt}
  \renewcommand{\arraystretch}{1.08}
  \begin{tabular}{p{0.18\linewidth}p{0.42\linewidth}p{0.32\linewidth}}
    \toprule
    Capability & Calls & Effect \\
    \midrule
    Geometry
      & \texttt{add\_box}, \texttt{add\_sphere}, \texttt{add\_cylinder},
        \texttt{add\_capsule}, \texttt{add\_plane}
      & Register fixed or free rigid bodies and visual/material attributes. \\
    Initial state
      & \texttt{set\_velocity}, \texttt{set\_rotation}
      & Set linear/angular velocity or a normalized orientation quaternion. \\
    Constraints
      & \texttt{add\_hinge}, \texttt{add\_slide}, \texttt{add\_spring}
      & Add one-DOF joints or a damped spatial spring between existing bodies. \\
    Environment
      & \texttt{set\_gravity}, \texttt{set\_duration}, \texttt{add\_camera}
      & Set the execution horizon, gravity vector, and look-at camera. \\
    Finalization
      & \texttt{freeze}
      & Add default ground/camera if absent and snapshot a \texttt{SceneDef}. \\
    \bottomrule
  \end{tabular}
\end{table}

\paragraph{Hard validation envelope.}
Table~\ref{tab:physdsl-bounds} gives the actual validator bounds. Stage~2 uses
\texttt{continuous\_levels=True}: mass, friction, and restitution remain the
submitted real values inside their intervals rather than being snapped to the
Stage-1 ladders.

\begin{table}[h]
  \centering
  \caption{PhysDSL construction-time bounds. Intervals are closed unless an
  inequality is written explicitly.}
  \label{tab:physdsl-bounds}
  \scriptsize
  \setlength{\tabcolsep}{4pt}
  \renewcommand{\arraystretch}{1.10}
  \begin{tabular}{p{0.25\linewidth}p{0.67\linewidth}}
    \toprule
    Field & Enforced contract \\
    \midrule
    Names
      & Pattern \texttt{[a-z\_][a-z0-9\_]*}; one namespace across bodies,
        joints, springs, and cameras; \texttt{ground/world/sky} reserved
        (\texttt{ground} is allowed for the plane). \\
    Body centers and count
      & $|x|,|y|<1.5$ and $0<z<2.5$\,m; a plane may use $z=0$.
        At least one body after finalization and at most 20 dynamic bodies;
        fixed scene furniture does not consume the dynamic-body cap. \\
    Geometric size
      & Each non-plane radius/half-size/half-height in $[0.01,1.0]$\,m;
        each plane half-size is at least $0.01$\,m and has no API upper cap. \\
    Material
      & Mass $[0.01,100]$\,kg, sliding friction $[0,2]$, restitution $[0,1]$,
        and every RGBA channel $[0,1]$. \\
    Initial state
      & Each linear and angular velocity component has magnitude at most
        $20$; orientation is a nonzero quaternion and is normalized. \\
    Joint
      & Nonzero axis; optional limits satisfy $\ell<u$; parent and child
        differ; only the parent may be \texttt{world}; one explicit joint per
        child. \\
    Spring
      & Stiffness $[0.1,10^4]$, damping $[0,10^3]$, body-local anchor
        components in $[-2,2]$\,m, and optional rest length $>0$; at least one
        endpoint is dynamic and neither endpoint is a plane. \\
    Environment
      & Duration $[0.1,60]$\,s; finite three-vector gravity; camera position
        and target must differ. \\
    \bottomrule
  \end{tabular}
\end{table}

\subsection{Reported evidence schemas}
\label{app:spec}

A reasoning intent specifies the semantic relation to probe and the evidence
needed to decide it. PhysDSL can express more worlds than the reported run
uses; the run enables exactly the three schemas in
Table~\ref{tab:manifest-schemas}, which jointly serve the eight active
families in Section~\ref{app:families}. The executor resolves every declared
role against the frozen world before constructing an item.

\begin{table}[t]
  \centering
  \caption{The three \texttt{SCENE\_MANIFEST} contracts enabled in the
  reported run. Braces denote a body-role mapping, not model-predicted state.}
  \label{tab:manifest-schemas}
  \scriptsize
  \setlength{\tabcolsep}{3.5pt}
  \renewcommand{\arraystretch}{1.12}
  \begin{tabular}{p{0.17\linewidth}p{0.43\linewidth}p{0.32\linewidth}}
    \toprule
    World class & Required role fields & Served active families \\
    \midrule
    Throwing
      & \texttt{balls}, three or more fixed \texttt{platforms}, two or more
        removable \texttt{walls}, \texttt{region\_kind}, and three
        destination regions derived from platform gaps
      & final destination; current-run judgment; counterfactual destination;
        inverse intervention \\
    Collision
      & dynamic \texttt{target}; \texttt{trigger}; exactly four dynamic
        \texttt{movers}; \texttt{relation\_pair=\{a,b,expected\}}
      & final state; final relation; motion attribution \\
    Motion
      & dynamic \texttt{motion\_target}; optional
        \texttt{speed\_pair=\{a,b\}}
      & motion-state change, including forward, speed-profile, and pairwise
        surface forms \\
    \bottomrule
  \end{tabular}
\end{table}

\section{Auditable World-Program Gallery}
\label{app:gallery}

Three human-authored \emph{reference worlds} illustrate the executable
world-program interface: Throwing, Collision, and Motion. We execute the
v01 programs with seed 7 in MuJoCo 3.6.0, retaining their colors, materials,
and lighting. Initialization images precede the first simulation step;
subsequent frames show the accepted world's execution. Program copies,
evidence schemas, accepted scenes, and execution records are stored under
\path{paper/assets/world_program_demo/}. The rendered components are stored
separately under \path{paper/figures/world_program_demo/}.

\subsection{Throwing: one program, four executable inputs}

\begin{figure}[H]
  \centering
  \input{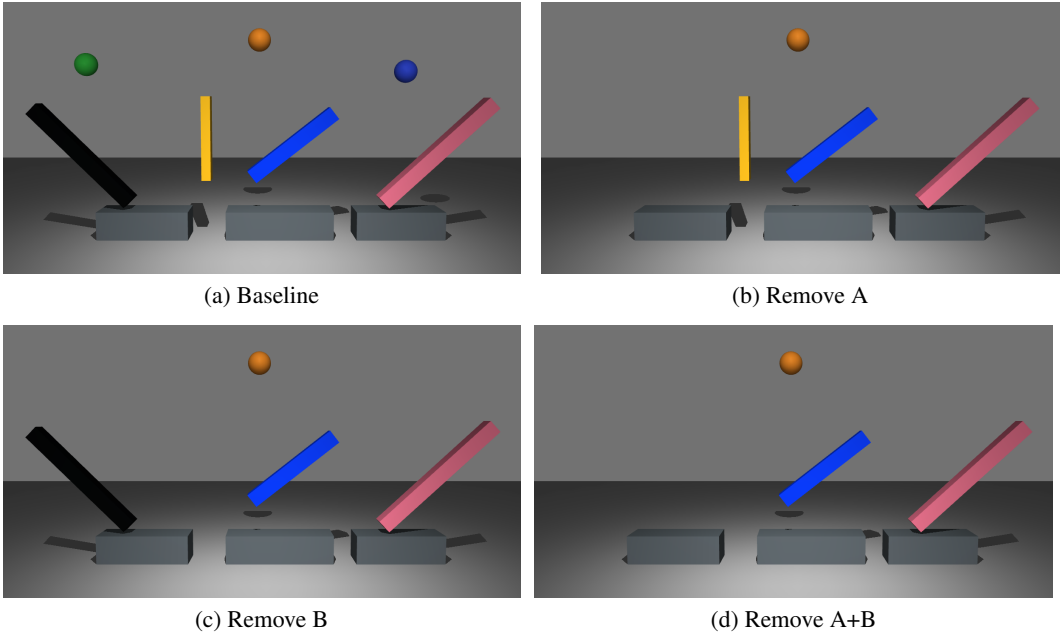}
  \caption{Throwing reference world and executable interventions. The frozen
  baseline and three altered inputs share the same camera and initial
  conditions for retained bodies. Each altered input removes the designated
  wall set and the non-query balls. The executor runs these inputs separately
  to obtain the region readouts used for question and answer construction.}
  \label{fig:throwing-program-gallery}
\end{figure}

The reference program defines three balls, three platforms, four walls, and
three destination regions. Figure~\ref{fig:throwing-program-gallery} uses the
orange ball (\texttt{ball\_b}) as the query. Wall A is
\texttt{black\_left\_wall}; wall B is \texttt{orange\_wall}.
Each branch preserves the retained bodies' initial conditions and removes
the designated walls together with the two non-query balls.

The orange ball settles in the left pit in the baseline and in the middle pit
when wall A and the other balls are removed. This executed difference supplies
the counterfactual answer. The four evidence objects in
Table~\ref{tab:throwing-branch-inputs} also support inverse questions by
comparing which interventions reach a requested destination.

\begin{table}[h]
  \centering\small
  \caption{Throwing evidence objects. Each altered input is executed with the
  same seed and horizon as the frozen baseline.}
  \label{tab:throwing-branch-inputs}
  \setlength{\tabcolsep}{5pt}
  \begin{tabular}{lll}
    \toprule
    Evidence key & Elided bodies & Readout \\
    \midrule
    $\tau_0(b)$ & none & baseline settled region \\
    $\tau_A(b)$ & wall A and all non-query balls & altered settled region \\
    $\tau_B(b)$ & wall B and all non-query balls & altered settled region \\
    $\tau_{AB}(b)$ & walls A+B and all non-query balls & altered settled region \\
    \bottomrule
  \end{tabular}
\end{table}

\newpage
\begingroup
\setlength{\intextsep}{6pt}
\subsection{Collision: roles and contact evidence}

The evidence schema names \texttt{swing\_striker} as the trigger,
\texttt{blue\_plate} as the target, four attribution candidates, and the
relation pair \texttt{orange\_ball}/\texttt{blue\_plate}. These roles support
attribution, final-state, and relation questions.

\begin{figure}[H]
  \centering
  \input{./figures/world_program_demo/collision/panels.tex}
  \caption{Collision reference world. The initialization and later frames
  connect the entity roles to execution. The lower panels bracket contact
  between the orange ball and blue plate; times are measured from simulation start.}
  \label{fig:collision-motion-gallery}
\end{figure}

\subsection{Motion: an apparatus for motion and contact}

The Motion world combines a ramp, a platform with a separate surface patch,
a slider, and two balls. The green ball descends the ramp and contacts the
amber slider, which subsequently contacts the blue ball. Its trajectory
provides evidence for the active motion-state family.

\begin{figure}[H]
  \centering
  \input{./figures/world_program_demo/motion/panels.tex}
  \caption{Motion reference world. Initialization and two execution states
  show the ramp-to-platform motion and resulting contacts. All frames share
  the same camera; times are measured from simulation start.}
  \label{fig:motion-program-gallery}
\end{figure}

\endgroup

\section{Intervention-Grounded Task Contracts}
\label{app:families}

The implementation follows one order for every family: define the physical
semantic, test whether the executed evidence makes it decidable, map the
admitted case to an answer semantic, and only then realize a question and
option layout. Tables~\ref{tab:throwing-family-contracts} and
\ref{tab:other-family-contracts} enumerate all eight active families. Frozen
registry entries are not counted.

\begin{table}[t]
  \centering
  \caption{Exact contracts for the four active throwing families. The option
  descriptions give canonical semantics; an eligible surface layout may
  permute slots and substitute compatible distractors, but must preserve a
  unique binding for the admitted answer semantic.}
  \label{tab:throwing-family-contracts}
  \scriptsize
  \setlength{\tabcolsep}{3pt}
  \renewcommand{\arraystretch}{1.12}
  \begin{tabular}{p{0.20\linewidth}p{0.35\linewidth}p{0.37\linewidth}}
    \toprule
    Family & Representative item and option semantics & Evidence, binding,
    and suppression \\
    \midrule
    \texttt{throwing\_\allowbreak final\_\allowbreak destination}
      & ``What is the final destination of the [referent]?'' Regions use
        \{left pit, right pit, center pit, none\}; an unsettled layout uses
        \{left, right, neither, none\}.
      & Baseline only. A region requires final speed $\leq0.30$\,m/s, a
        non-ejected status, exactly one derived gap interval, and final height
        no more than $0.30$\,m above the platform tops. A settled region binds
        that region; otherwise \emph{neither} is bound rather than guessed. \\
    \texttt{throwing\_\allowbreak current\_\allowbreak run\_\allowbreak judgment}
      & ``Will the [referent] eventually drop into the left pit?'' Options are
        \{Yes, No, Can not answer, Uncertain\}; a second surface asks whether
        it reaches a different destination region than a named one.
      & Baseline only. \emph{Yes} iff the settled-region readout equals the
        named target; otherwise \emph{No}, including an unsettled/ejected
        readout. Target choice alternates between matching and nonmatching
        regions across ball indices. \\
    \texttt{counterfactual\_\allowbreak destination}
      & ``If we removed the [wall] and other balls, which pit would the
        [referent] drop into?'' Options are \{left, right, center, none\}.
      & Matched baseline and one elision branch. Admit only when the branch is
        available, settles in one region, and differs from the baseline
        region. The branch region is the answer; failed, unsettled, or
        unchanged branches suppress the item. \\
    \texttt{inverse\_\allowbreak intervention}
      & ``What can we do to make the [referent] drop into the [target] pit?''
        Options are \{remove A + other balls, remove B + other balls, remove
        both + other balls, none\}.
      & Baseline plus all three A/B/A+B branches. In a first fixed-order pass,
        the first non-baseline target reached by exactly one branch binds that
        intervention. Only if no target has a unique achiever, a second pass
        binds \emph{none} to the first non-baseline target reached by no branch.
        A multiply achieved target is skipped; missing branches suppress. \\
    \bottomrule
  \end{tabular}
\end{table}

\begin{table}[t]
  \centering
  \caption{Exact contracts for the other four active families. Canonical
  options are shown; eligible layouts may change slot order and compatible
  distractors while preserving the admitted answer semantic.}
  \label{tab:other-family-contracts}
  \scriptsize
  \setlength{\tabcolsep}{3pt}
  \renewcommand{\arraystretch}{1.12}
  \begin{tabular}{p{0.20\linewidth}p{0.35\linewidth}p{0.37\linewidth}}
    \toprule
    Family & Representative item and option semantics & Evidence, binding,
    and suppression \\
    \midrule
    \texttt{collision\_\allowbreak final\_state}
      & ``Which phrase best describes the final pose of the [target]?''
        Options are \{standing upright, leaning, lying horizontally,
        floating\}.
      & Terminal status and speed. Ejected with $z>0.35$ binds floating;
        fallen binds lying; moving with speed $\geq0.10$ binds leaning; stable
        with speed $\leq0.20$ binds standing. Other/borderline combinations
        suppress the item. \\
    \texttt{collision\_\allowbreak final\_\allowbreak relation}
      & ``Which phrase best describes the interaction between A and B during the run?'' Options are
        \{ends in contact, never touches, passes near without touching, ends
        vertically separated without contact\}; a binary surface uses
        \{Yes, No, Sometimes, Uncertain\}.
      & Contact is trajectory-level: any recorded pairwise contact dominates
        and binds the contact semantic (canonically rendered as ``ends in
        contact''). Otherwise vertical separation requires final
        $|\Delta z|\geq0.30$ and lateral gap $<0.30$\,m; pass-by requires
        initial or final distance $\leq0.20$\,m; the remainder is no-contact.
        The computed class must equal the manifest declaration or the family
        is suppressed. \\
    \texttt{motion\_attribution}
      & ``Which object experienced a change in motion as a result of the
        [trigger]?'' The four options are the manifest's four mover bodies.
      & Candidates must be distinct, dynamic, and different from the trigger.
        Eligibility requires trigger contact and velocity-change magnitude
        $\lVert\mathbf v_f-\mathbf v_0\rVert_2>0.30$\,m/s; the argmax is bound
        only if its lead over a second eligible candidate is at least
        $0.30$\,m/s. The absence of an eligible candidate, or a lead below this
        margin, suppresses the item. \\
    \texttt{motion\_state\_change}
      & Three surfaces: motion change \{stop, slow but moving, speed up, same\};
        speed profile \{accelerate, decelerate, accelerate then decelerate,
        constant\}; pairwise semantics \{final tie, final A faster, final B
        faster, non-tied order reversal\}. The canonical middle options use
        ``consistently faster'' wording.
      & Require target initial speed $\geq0.10$\,m/s. Stop uses final speed
        $<0.10$; speed-change/profile boundaries use an absolute $0.30$\,m/s
        gap; pairwise order uses $0.10$\,m/s. Crossover is bound only when both
        endpoint orders are non-tied and reverse; otherwise the final order is
        bound. The pairwise surface requires both declared bodies. \\
    \bottomrule
  \end{tabular}
\end{table}

\subsection{Branch matrix and unique-answer binding}

For a query ball $b$, let $r_0,r_A,r_B,r_{AB}\in
\{\text{left},\text{right},\text{middle},\bot\}$ be the settled-region
readouts of the four executions in Table~\ref{tab:throwing-branch-inputs};
$\bot$ denotes no admitted region. Table~\ref{tab:branch-result-contract}
formalizes which cells each family must observe before any option letters are
assigned.

\begin{table}[t]
  \centering
  \caption{Branch-result contract used for counterfactual and inverse items.}
  \label{tab:branch-result-contract}
  \small
  \setlength{\tabcolsep}{6pt}
  \begin{tabular}{lcccc}
    \toprule
    Execution & Baseline & Remove A & Remove B & Remove A+B \\
    \midrule
    Stored semantic & $r_0$ & $r_A$ & $r_B$ & $r_{AB}$ \\
    Required for counterfactual(A) & yes & yes & no & no \\
    Required for inverse & yes & yes & yes & yes \\
    \bottomrule
  \end{tabular}
\end{table}

A counterfactual(A) item is admitted exactly when
$r_A\neq\bot$ and $r_A\neq r_0$, then binds answer semantic $r_A$. For an
inverse target $t\neq r_0$, define
$H_t=\{A:r_A=t,\ B:r_B=t,\ AB:r_{AB}=t\}$.
The generator requires all three branch views and makes two fixed-order passes
over left/right/middle. The first returns the first target with $|H_t|=1$ and
binds its sole lever. Only when no target has a unique achiever does the second
return the first target with $|H_t|=0$ and bind \emph{none of the above}; a
target with $|H_t|>1$ is skipped in both passes. Thus ``none'' is an
executor-derived negative, not a catch-all for failed execution, and an
ambiguous target does not preclude a decidable target elsewhere.

\subsection{Surface realization boundary}

Only after an answer semantic is admitted does the task-interface library
select a compatible stem and four-option layout and bind the semantic to an
index. A canonical family layout is used whenever a sampled surface lacks
the required semantic. This ordering keeps the physical decision rule
executor-determined while allowing multiple linguistic realizations of the
same evidence.

\section{Verification, Repair, and Branch Execution}
\label{app:status}

\paragraph{Typed termination.}
Every proposal receives exactly one status. \texttt{FAILED\_SYNTAX},
\texttt{FAILED\_SEMANTIC}, and \texttt{FAILED\_RUNTIME} cover parsing, eager
DSL validation, and \texttt{build\_scene} execution. \texttt{FAILED\_COMPILE},
\texttt{FAILED\_OVERLAP}, and \texttt{FAILED\_NAN} cover simulator-model,
initial-interpenetration, and numerical failures. \texttt{FAILED\_TIMEOUT} and
\texttt{FAILED\_RENDER} cover bounded execution and rendering.
\texttt{FAILED\_VQA} means the manifest or generated item violates its
contract. \texttt{TRIVIAL} means maximum displacement is below $0.05$\,m and
there is no collision. Only \texttt{OK} worlds with admitted items enter
Solver groups.

\paragraph{Bounded repair.}
Repairs are recorded as an ordered trace. Minor repairs address supported
syntax/argument variants, manifest repairs recover fields mechanically
determined by authored geometry, and geometry repairs clamp workspace
positions or separate simple initial overlap. Repair does not add a new body,
joint, spring, or declared causal relation. The strongest applied tier yields
a proposal-credit multiplier of $0.90$ (minor), $0.80$ (manifest), or $0.75$
(geometry); unrepaired worlds use $1.0$.

\paragraph{Frozen matched branches.}
After repair, the executor calls \texttt{freeze()} once. A branch clones that
immutable \texttt{SceneDef}, elides its named body set and every incident
joint/spring, and re-executes with the baseline seed and horizon. Identical
removal sets share one cached execution. At most six distinct removal sets are
run per world; failed/capped branches are absent, causing dependent items to be
suppressed. The throwing scheduler places the A/B/A+B inverse sets for the
first two balls before compatible counterfactual requests, so the latter can
reuse the same branch views.

\begin{algorithm}[t]
  \caption{Executor-determined item construction under an explicit contract}
  \label{alg:item-construction}
  \begin{algorithmic}[1]
    \Require proposal module $p$, stage-2 DSL, simulation seed $s$
    \State parse \texttt{build\_scene} and literal \texttt{SCENE\_MANIFEST}
    \State execute the sandboxed builder with eager DSL checks
    \State apply bounded repair and freeze one immutable world $W$
    \State compile, execute, and render baseline $\tau_0\leftarrow\mathcal{E}(W,s)$
    \State validate manifest roles against $W$ and enumerate required elisions
    \For{each distinct removal set $B$ up to the branch cap}
      \State $W_B\leftarrow W$ with $B$ and incident constraints elided
      \State $\tau_B\leftarrow\mathcal{E}(W_B,s)$; cache by $B$
    \EndFor
    \For{each requested active family $f$}
      \State compute evidence predicates from $\tau_0$ and required $\tau_B$
      \If{all decidability and margin conditions for $f$ hold}
        \State derive answer semantic; select a compatible surface layout
        \State bind semantic to one option index and emit the item
      \EndIf
    \EndFor
  \end{algorithmic}
\end{algorithm}

\section{Optimization and Active Curriculum}
\label{app:reward}

\paragraph{General objective.}
Let $z$ be the terminal proposal status, $n$ the number of retained Solver
rollouts, $G_{\min}=4$, and $I$ the intrinsic score. Before repair, anti-copy,
saturation control, and family scaling, the implementation defines
\begin{equation}
  r_{\mathrm{base}}=
  \begin{cases}
    \lambda_{\mathrm{fail}}c_z, & z\neq\texttt{OK},\\
    0, & z=\texttt{OK}\ \text{and}\ n<G_{\min},\\
    \lambda_{\mathrm{learn}}^{\mathrm{eff}}r_{\mathrm{learn}}
      +\lambda_{\mathrm{int}}I, & \text{otherwise}.
  \end{cases}
  \label{eq:proposal-reward-appendix}
\end{equation}
Equation~\ref{eq:proposal-reward-appendix} separates terminal failure,
insufficient retained evidence, and admitted worlds. Here
$r_{\mathrm{learn}}=1-2|\bar a_{\mathrm{prop}}-\tfrac{1}{2}|$ is the
instance-level tent statistic, where $\bar a_{\mathrm{prop}}$ denotes the
retained-rollout Solver accuracy used to score the proposal. The Solver
receives exact option-match reward. Status prices are $c_z=-1.0$ for syntax,
semantic, and VQA failure,
$-0.8$ for runtime failure, $-0.5$ for compile and overlap failure, $-0.3$
for numerical, timeout, and render failure, and $0$ for a trivial world.

\paragraph{Reported instantiation and hierarchical co-evolution.}
Co-evolution operates through a two-level curriculum. At the macro level,
family sampling tracks the Solver's competence boundary across reasoning
intents: after an eight-step prior-weighted cold start, selection uses
$\max(0.05,4\bar a_f(1-\bar a_f))$ multiplied by the structural priors in
Table~\ref{tab:reward-config}, so saturated or unsolvable families are
down-weighted. The family-difficulty multiplier further scales Proposer
credit by rolling $\bar a_f$, so Solver outcomes reshape both which intents
are sampled and how admitted worlds are valued. At the micro level,
conditioned on the sampled intent $\sigma$, the shared policy explores the
program space for valid, physically non-trivial environments. Instance-level
scoring prioritizes physical validity, collision feasibility, and structural
diversity $I$, with the instance-level tent coefficient held at
$\lambda_{\mathrm{learn}}^{\mathrm{eff}}=0$. This hierarchy combines
competence-adaptive task selection with verifier-grounded exploration of
executable worlds. The executor supplies the verifiable answers that connect
both levels. Executable
demonstrations remain available as syntactic references. Through this
dual-level loop, Solver progression reshapes the training distribution via the
adaptive intent curriculum, and shared-parameter updates of $\pi_\theta$
align world generation with physical reasoning.

The active proposal signals are typed failure/validity, repair credit,
structural diversity, anti-copy, saturation control, and family-difficulty
scaling; the active Solver signal is exact option match.
Table~\ref{tab:reward-config} records these controls.

\begin{table}[t]
  \centering
  \caption{Reward and sampling controls in the reported 100-step run.}
  \label{tab:reward-config}
  \small
  \setlength{\tabcolsep}{5pt}
  \renewcommand{\arraystretch}{1.08}
  \begin{tabular}{p{0.43\linewidth}p{0.49\linewidth}}
    \toprule
    Component & Reported configuration \\
    \midrule
    Failure scale $\lambda_{\mathrm{fail}}$ & $1.5$ \\
    Instance-level tent coefficient
      $\lambda_{\mathrm{learn}}^{\mathrm{eff}}$
      & held at $0$ (validity/diversity dominate instance scoring) \\
    Intrinsic weight $\lambda_{\mathrm{int}}$ & $0.1$; structural diversity only \\
    Repair factors & $0.90$ minor; $0.80$ schema; $0.75$ geometry \\
    Near-copy control & similarity $\geq0.95$ capped at $0.1$ before repair and
      family scaling \\
    Exact/within-group duplicate proposal & pre-family reward override $-0.8$ \\
    Edit-distance diversity weight & $1.5$ \\
    Saturation control & threshold $0.75$; weight $1.0$ \\
    Family-difficulty multiplier & $\operatorname{clip}(1+(0.5-\bar a_f),0.5,1.5)$ \\
    Rolling family statistic & 64 most recent per-world family accuracies \\
    Family-sampling cold start & 8 rounds \\
    Family-sampling priors
      & $2.0\times$ for priority families; $2.0\times$ for
        \texttt{collision\_\allowbreak final\_\allowbreak relation}; $1.2\times$ for each branch family;
        $1.0\times$ otherwise \\
    Solver reward scale & $1.5$ \\
    \bottomrule
  \end{tabular}
\end{table}

Exact Solver outcomes update the rolling family statistics used for
subsequent rounds. Family sampling directs task generation toward the
Solver's competence boundary, and the family-difficulty multiplier adjusts
Proposer credit across intents. Within each intent, validity, repair,
diversity, and saturation controls shape executable world generation.

\section{Reproducibility, Training, and Evaluation Configuration}
\label{app:hparams}

\subsection{Role contracts and update loop}

The shared policy alternates between two tagged interfaces. The Proposer
receives a reasoning intent and returns a world program plus manifest inside
\texttt{<propose>} tags. The Solver receives a rendered video and an
executor-authored multiple-choice item and returns one option inside
\texttt{<answer>} tags; it has no simulator tool. The fixed executor joins the
roles by validating programs, executing the required baseline/branches, and
binding each admitted answer.

\begin{algorithm}[t]
  \caption{Reported verifier-grounded shared-policy training loop}
  \label{alg:loop}
  \begin{algorithmic}[1]
    \Require shared policy $\pi_\theta$, executor $\mathcal{E}$,
      intent curriculum $\mathcal{C}$, intent batch $N$, group size $G$
    \For{$t=1,\ldots,100$}
      \State sample $\{\sigma_n\}_{n=1}^{N}\sim\mathcal{C}$ with scene-kind minima
      \State sample $G$ program/manifest proposals for each $\sigma_n$
      \State validate proposals and execute Algorithm~\ref{alg:item-construction}
      \State retain up to the per-scene and per-step Solver-group caps
      \State sample $G$ Solver answers per retained item; score exact option match
      \State compute proposal rewards and role-wise GRPO advantages
      \State update the shared $\pi_\theta$ from Proposer and Solver rows
      \State update rolling family accuracy and $\mathcal{C}$
    \EndFor
  \end{algorithmic}
\end{algorithm}

\subsection{Resolved training configuration}

Table~\ref{tab:training-config} transcribes the v7i launcher and current
configuration builder. The launcher fixes a 100-step horizon and saves every
25 steps. We report the terminal step-100 checkpoint and show every scheduled
intermediate checkpoint in Figure~\ref{fig:checkpoint-dynamics}.

\begin{table}[t]
  \centering
  \caption{Resolved configuration for the reported \physevolve{} run.}
  \label{tab:training-config}
  \scriptsize
  \setlength{\tabcolsep}{4pt}
  \renewcommand{\arraystretch}{1.10}
  \begin{tabular}{p{0.31\linewidth}p{0.61\linewidth}}
    \toprule
    Component & Configuration \\
    \midrule
    Base and horizon
      & Qwen3-VL-8B-Instruct; 100 steps; checkpoints at 25/50/75/100 \\
    Run multiplicity
      & One training run per reported configuration; no training-seed sweep \\
    Compute
      & 4 GPUs; 32 Ray CPUs; 4 executor workers \\
    Proposer sampling
      & $N=4$ intent groups per step; $G=4$ programs per intent; at least one
        throwing, collision, and motion intent per step \\
    Solver retention and rollout
      & At most 5 items per scene and 32 groups per step; 4 answers per group;
        rollout chunk size 8 \\
    Visual variants
      & One additional color permutation per accepted baseline world; one
        executor camera; 32 rendered frames at $480\!\times\!480$ \\
    Golden demonstration pool
      & 10 throwing, 8 collision, and 6 motion programs; active for steps
        1--100 \\
    Auxiliary geometry item
      & Camera-depth bypass enabled; at most one executor-geometry item on an
        eligible collision/motion scene, outside the eight sampled intents \\
    Generation
      & temperature $0.7$; top-$p=0.95$; top-$k=20$; 6,144-token prompt and
        4,096-token response budgets \\
    GRPO
      & 1 epoch; clip ratio $0.2$; learning rate $10^{-6}$; entropy coefficient
        $0$ \\
    Batch sizes
      & training batch 4; prompt-group mini-batch 1; per-GPU micro-batch 1 \\
    Reference regularization
      & frozen reference policy; low-variance actor KL and reward KL coefficient
        $0.03$ \\
    Reference buffer
      & enabled during demonstrations; capacity 32; 2 references per prompt \\
    Family sampling
      & 8-step prior-weighted cold start; accuracy-adaptive sampling thereafter.
        Structural priors throughout: $2.0\times$ for priority and relation
        coverage; $1.2\times$ for counterfactual/inverse families. \\
    Online construction seeds
      & executor sequence starts at 9001; curriculum/reference RNG base 9001
        (resolved code defaults) \\
    Executor limits
      & $0.002$\,s timestep; 30\,s simulation and 15\,s render timeouts; at
        most 6 distinct branch executions per world \\
    \bottomrule
  \end{tabular}
\end{table}

\subsection{Evaluation protocols}
\label{app:evaluation-protocols}

Table~\ref{tab:evaluation-protocols} gives the evaluation contract used for
each benchmark.

\begin{table}[t]
  \centering
  \caption{Evaluation protocols for the three reported benchmarks.}
  \label{tab:evaluation-protocols}
  \scriptsize
  \setlength{\tabcolsep}{3pt}
  \renewcommand{\arraystretch}{1.12}
  \begin{tabular}{p{0.14\linewidth}p{0.12\linewidth}p{0.18\linewidth}p{0.18\linewidth}p{0.28\linewidth}}
    \toprule
    Benchmark & Items & Visual input & Generation & Scoring \\
    \midrule
    PhysBench
      & 9,802
      & 8 uniform frames
      & greedy; 2,048 tokens
      & official prompts, normalization, and scorer \\
    CausalPhys
      & 3,057
      & 5 uniform frames
      & CoT; greedy; 512 tokens
      & final A--D answer; unparsed output is incorrect \\
    ContPhy
      & 1,950
      & 10 uniform frames
      & greedy; 64 tokens
      & normalized MC or normalized open-response exact match \\
    \bottomrule
  \end{tabular}
\end{table}

PhysBench frames are passed directly to the Qwen processor as an array. For
CausalPhys, five annotations whose gold field is not an A--D label are
excluded before scoring. ContPhy evaluates one question per generation and
normalizes extracted open responses by lowercasing and stripping whitespace.
All models within one benchmark share the item list, visual preprocessing,
prompt, decoding, answer parser, and scorer. An unparsed response is scored as
incorrect.

We report accuracy and percentage-point differences on the shared item set
within each benchmark.

\section{Extended Results}
\label{app:physbench-breakdown}

\subsection{PhysBench taxonomy}

Table~\ref{tab:physbench-aligned} reports the matched claim-facing views.
Table
\ref{tab:physbench-complete-taxonomy} then gives every top-level task and
ability partition from the same step-100 evaluation, plus the three physical
dynamics subtypes used in the main text.

\begin{table}[t]
  \centering
  \caption{PhysBench accuracy and paired differences for matched models.}
  \label{tab:physbench-aligned}
  \scriptsize
  \setlength{\tabcolsep}{2.8pt}
  \resizebox{\linewidth}{!}{%
  \begin{tabular}{lccccc}
    \toprule
    View ($n$)
    & Base
    & Naive
    & $\Delta$ vs.\ Base
    & \physevolve{}
    & $\Delta$ vs.\ Base \\
    \midrule
    Overall (9,802)
    & 52.92 & 52.39 & $-0.53$
    & 53.47 & $+0.55$ \\
    Dynamics (2,932)
    & 44.37 & 44.44 & $+0.07$
    & 45.29 & $+0.92$ \\
    Throwing (410)
    & 46.10 & 45.85 & $-0.24$
    & 54.88 & $+8.78$ \\
    Collision (650)
    & 38.62 & 38.77 & $+0.15$
    & 40.77 & $+2.15$ \\
    Motion (1,433)
    & 54.71 & 55.48 & $+0.77$
    & 56.45 & $+1.74$ \\
    Prediction (1,430)
    & 42.17 & 42.87 & $+0.70$
    & 43.71 & $+1.54$ \\
    Judgment (767)
    & 34.94 & 35.20 & $+0.26$
    & 40.03 & $+5.08$ \\
    Reasoning (2,056)
    & 43.09 & 40.42 & $-2.68$
    & 43.92 & $+0.83$ \\
    Static (1,899)
    & 60.66 & 61.40 & $+0.74$
    & 62.24 & $+1.58$ \\
    \bottomrule
  \end{tabular}%
  }
  \vspace{2pt}

  \begin{minipage}{0.98\linewidth}
    \footnotesize
    Accuracies and differences are percentage points. Difference columns
    compare Naive and \physevolve{} with Base. Task, subtype, and ability
    levels overlap; rows within each top-level partition do not.
  \end{minipage}
\end{table}

\begin{table}[t]
  \centering
  \caption{Complete top-level PhysBench task and ability partitions for the
  matched Base and \physevolve{} models. The dynamics-subtype rows are nested
  inside Dynamics and therefore do not form a complete partition.}
  \label{tab:physbench-complete-taxonomy}
  \scriptsize
  \setlength{\tabcolsep}{5pt}
  \renewcommand{\arraystretch}{1.05}
  \begin{tabular}{llrrr}
    \toprule
    Taxonomy level & View ($n$) & Base & \physevolve{} & $\Delta$ \\
    \midrule
    Task & Dynamics (2,932) & 44.37 & 45.29 & +0.92 \\
         & Relationships (2,088) & 61.97 & 63.46 & +1.48 \\
         & Scene (2,240) & 45.13 & 45.62 & +0.49 \\
         & Property (2,542) & 62.20 & 61.61 & $-0.59$ \\
    \midrule
    Dynamics subtype & Throwing (410) & 46.10 & 54.88 & +8.78 \\
                     & Collision (650) & 38.62 & 40.77 & +2.15 \\
                     & Motion (1,433) & 54.71 & 56.45 & +1.74 \\
    \midrule
    Ability & Perception (919) & 60.39 & 56.04 & $-4.35$ \\
            & Dynamic (189) & 75.13 & 75.66 & +0.53 \\
            & Static (1,899) & 60.66 & 62.24 & +1.58 \\
            & Comparison (1,423) & 45.89 & 44.91 & $-0.98$ \\
            & Reasoning (2,056) & 43.09 & 43.92 & +0.83 \\
            & Identify (1,119) & 82.93 & 82.84 & $-0.09$ \\
            & Judgment (767) & 34.94 & 40.03 & +5.08 \\
            & Prediction (1,430) & 42.17 & 43.71 & +1.54 \\
    \bottomrule
  \end{tabular}
\end{table}

\subsection{Complete contextual reference table}

Table~\ref{tab:full-reference-results} restores the complete set of views for
the open-weight contextual references. These heterogeneous scales are not
used for component attribution; Base, Naive, and \physevolve{} remain the
matched 8B comparison.

\begin{table}[t]
  \centering
  \caption{Full accuracy (\%) table for open-weight context and matched 8B
  models.}
  \label{tab:full-reference-results}
  \scriptsize
  \setlength{\tabcolsep}{1.7pt}
  \renewcommand{\arraystretch}{1.05}
  \resizebox{\linewidth}{!}{%
  \begin{tabular}{l*{8}{c}@{\hspace{4pt}}*{3}{c}@{\hspace{4pt}}*{3}{c}}
    \toprule
    & \multicolumn{8}{c}{PhysBench}
    & \multicolumn{3}{c}{CausalPhys}
    & \multicolumn{3}{c}{ContPhy} \\
    \cmidrule(lr){2-9}\cmidrule(lr){10-12}\cmidrule(l){13-15}
    Model & Ovr. & Dyn. & Throw. & Coll. & Motion & Pred. & Judg. & Reason.
      & Ovr. & Antic. & Interv. & Ovr. & CF-Rm. & Goal-Pit \\
    \midrule
    InternVL2.5-8B
      & 43.8 & 42.1 & 40.0 & 38.9 & 27.7 & 43.0 & 32.7 & 27.4
      & 50.2 & 51.5 & 42.4 & 31.4 & 48.4 & 23.3 \\
    Llama-3.2-11B-Vision
      & 38.7 & 43.2 & 36.6 & 35.5 & 27.2 & 44.1 & 30.8 & 29.6
      & 44.2 & 44.1 & 40.1 & 20.7 & 0.0 & 20.0 \\
    Pixtral-12B
      & 46.9 & 44.6 & 41.2 & 39.2 & 35.2 & 46.9 & 31.0 & 35.1
      & 50.2 & 53.0 & 44.5 & 21.5 & 40.2 & 13.3 \\
    Qwen2.5-VL-7B
      & 49.4 & 44.3 & 44.4 & 36.5 & 50.6 & 44.1 & 33.1 & 36.5
      & 52.6 & 56.5 & 45.7 & 25.9 & 11.5 & 24.4 \\
    Qwen3-VL-32B
      & 60.9 & 48.7 & 47.1 & 46.0 & 83.5 & 48.0 & 44.3 & 51.4
      & 58.9 & 61.9 & 52.9 & 41.0 & 45.9 & 26.7 \\
    \midrule
    Qwen3-VL-8B Base
      & 52.9 & 44.4 & 46.1 & 38.6 & 54.7 & 42.2 & 34.9 & 43.1
      & 53.9 & 55.9 & 42.8 & 37.1 & 44.3 & 20.0 \\
    Naive self-play
      & 52.4 & 44.4 & 45.9 & 38.8 & 55.5 & 42.9 & 35.2 & 40.4
      & 55.8 & 57.6 & 45.9 & 37.4 & 43.4 & 25.6 \\
    \physevolve{}
      & 53.5 & 45.3 & 54.9 & 40.8 & 56.5 & 43.7 & 40.0 & 43.9
      & 54.5 & 58.4 & 45.0 & 39.5 & 51.6 & 44.4 \\
    \bottomrule
  \end{tabular}%
  }
\end{table}

\subsection{CausalPhys categories and ContPhy domains}

Table~\ref{tab:cross-benchmark-extended} exposes all four CausalPhys
categories and all four ContPhy physical domains from the reported protocols.
The intervention-operation views used in the main text are included after the
domain partition.

\begin{table}[t]
  \centering
  \caption{Matched cross-benchmark accuracy and complete category/domain
  breakdowns.}
  \label{tab:cross-benchmark-extended}
  \scriptsize
  \setlength{\tabcolsep}{4pt}
  \renewcommand{\arraystretch}{1.10}
  \begin{tabular}{lccc}
    \toprule
    Benchmark view ($n$) & Base & \physevolve{} & $\Delta$ \\
    \midrule
    CausalPhys Overall (3,057)
      & 53.94 & 54.50 & $+0.56$ \\
    \quad Anticipation (896)
      & 55.92 & 58.37 & $+2.46$ \\
    \quad Goal orientation (637)
      & 50.71 & 49.61 & $-1.10$ \\
    \quad Intervention (573)
      & 42.76 & 45.03 & $+2.27$ \\
    \quad Perception (951)
      & 60.99 & 59.83 & $-1.16$ \\
    \midrule
    ContPhy Overall (1,950)
      & 37.08 & 39.54 & $+2.46$ \\
    \quad Cloth (450)
      & 48.00 & 47.11 & $-0.89$ \\
    \quad Fluid (450)
      & 24.89 & 28.44 & $+3.55$ \\
    \quad Rope (600)
      & 30.83 & 32.17 & $+1.33$ \\
    \quad Soft body (450)
      & 46.67 & 52.89 & $+6.22$ \\
    \midrule
    \quad Counterfactual removal (122)
      & 44.26 & 51.64 & $+7.38$ \\
    \quad Goal-Pit (90)
      & 20.00 & 44.44 & $+24.44$ \\
    \bottomrule
  \end{tabular}
\end{table}

The complete branch-path control is reported in Table~\ref{tab:ablations}.

\subsection{Joint Proposer--Solver updates}

An additional run removed Proposer-role rows from the shared-policy update
while retaining Proposer generation, the executor, and the family set. Its
training answer parse rate remains valid through step 64 and becomes zero for
steps 65--100. Table~\ref{tab:joint-role-updates} reports endpoint accuracy
and answer parse rates for the joint and Solver-only update configurations.

\subsection{Task-level transfer and paired answer changes}
\label{app:transfer-analysis}

Table~\ref{tab:causalphys-fine-grained} gives the exact accuracies underlying
Figure~\ref{fig:transfer-task-differences}. The CausalPhys Naive and
branch-disabled results use the same video-packing protocol as the Base
and full-model evaluations. All comparisons retain the same
valid-gold evaluation set.

\begin{table}[!htbp]
\centering\small
\caption{Complete accuracies for the task-level transfer comparison (\%). CausalPhys controls use the matched video-packing protocol.}
\label{tab:causalphys-fine-grained}
\setlength{\tabcolsep}{4pt}
\begin{tabular*}{\linewidth}{@{\extracolsep{\fill}}lrrrrr@{}}
\toprule
Task & $n$ & Base & Naive & w/o branch & \physevolve{} \\
\midrule
\multicolumn{6}{l}{\emph{ContPhy}} \\
Counterfactual Removal & 122 & 44.26 & 43.44 & 43.44 & \textbf{51.64} \\
Goal-Pit & 90 & 20.00 & 25.56 & 21.11 & \textbf{44.44} \\
\multicolumn{6}{l}{\emph{CausalPhys / Anticipation}} \\
Collision Prediction & 300 & 53.67 & 50.67 & \textbf{54.67} & 51.67 \\
Fluid Flow & 200 & 47.50 & \textbf{60.50} & 35.50 & 58.50 \\
Intention Speculation & 196 & 45.92 & \textbf{52.55} & 42.35 & 48.47 \\
Deformation & 200 & 77.50 & 70.00 & \textbf{78.00} & \textbf{78.00} \\
\multicolumn{6}{l}{\emph{CausalPhys / Intervention}} \\
Action Substitution & 99 & 38.38 & 46.46 & 44.44 & \textbf{47.47} \\
Spatial Manipulation & 151 & 43.05 & \textbf{46.36} & 41.06 & 45.03 \\
Temporal Shifting & 149 & 54.36 & \textbf{59.73} & 50.34 & 57.05 \\
Viewpoint Transformation & 174 & \textbf{35.06} & 33.33 & 33.33 & 33.33 \\
\bottomrule
\end{tabular*}
\end{table}

Table~\ref{tab:physbench-paired} uses the official PhysBench answer-matching
rule to partition the matched questions by
terminal-checkpoint correctness. The difference between the two discordant
counts, divided by the row's sample size, recovers the accuracy change.
Throwing and Motion are taxonomy subtypes; Judgment is an ability view and
can overlap the subtype views. Counts are therefore interpreted within each
row.

\begin{table}[!htbp]
\centering\small
\caption{Paired terminal-checkpoint correctness on PhysBench. Each row partitions the same evaluation items by Base and PhysEvolve correctness.}
\label{tab:physbench-paired}
\setlength{\tabcolsep}{4pt}
\begin{tabular}{lrrrrrr}
\toprule
Task & $n$ & Both correct & Ours only & Base only & Neither & $\Delta$ (pp) \\
\midrule
Overall & 9802 & 4487 & 754 & 700 & 3861 & +0.55 \\
Throwing & 410 & 156 & 69 & 33 & 152 & +8.78 \\
Judgment & 767 & 220 & 87 & 48 & 412 & +5.08 \\
Motion & 1433 & 730 & 79 & 54 & 570 & +1.74 \\
\bottomrule
\end{tabular}
\end{table}

\begin{figure}[!htbp]
  \centering
  \begin{minipage}{\linewidth}\raggedright
  \input{./figures/analysis_data/contphy_cases.tex}
  \end{minipage}
  \caption{Matched ContPhy examples for Counterfactual Removal and Goal-Pit.
  Each example displays three of the ten evaluation input frames, the original
  question and options, the extracted model answers, and the reference answer.
  The examples illustrate reasoning about removal and a requested destination
  in held-out soft-body scenes. The Goal-Pit reference answer is ``None of the
  above.''}
  \label{fig:contphy-transfer-examples}
\end{figure}

\end{document}